\documentclass[runningheads]{llncs}
\usepackage{graphicx}
\usepackage[linesnumbered,ruled]{algorithm2e}
\usepackage{amsmath,amssymb,amsfonts}
\usepackage{algorithmic}
\usepackage{graphicx}
\usepackage{textcomp}
\usepackage{xcolor}
\usepackage{bm}
\usepackage{multirow}
\usepackage{booktabs}
\usepackage[figuresright]{rotating}
\begin{document}
\title{Accelerating the Evolutionary Algorithms by Gaussian Process Regression with $\epsilon$-greedy acquisition function}
%
%
%
%
\titlerunning{Accelerating the Evolutionary Algorithms}  
%
\author{Rui Zhong\inst{1} \and Enzhi Zhang\inst{2} \and Masaharu Munetomo\inst{3}}
\authorrunning{Rui Zhong, Enzhi Zhang, Masaharu Munetomo}
\institute{Graduate School of Information Science and Technology, Hokkaido University, Sapporo, 060-0814, Japan,\\\email{rui.zhong.u5@elms.hokudai.ac.jp},\\ 
\and
Graduate School of Information Science and Technology, Hokkaido University, Sapporo, 060-0814, Japan,\\\email{enzhi.zhang.n6@elms.hokudai.ac.jp},\\ 
\and
Information Initiative Center, Hokkaido University, Sapporo 060-811, Japan, \\ \email{munetomo@iic.hokudai.ac.jp}}

\maketitle              
\begin{abstract}
In this paper, we propose a novel method to estimate the elite individual to accelerate the convergence of optimization. Inspired by the Bayesian Optimization Algorithm (BOA), the Gaussian Process Regression (GPR) is applied to approximate the fitness landscape of original problems based on every generation of optimization. And simple but efficient $\epsilon$-greedy acquisition function is employed to find a promising solution in the surrogate model. Proximity Optimal Principle (POP) states that well-performed solutions have a similar structure, and there is a high probability of better solutions existing around the elite individual. Based on this hypothesis, in each generation of optimization, we replace the worst individual in Evolutionary Algorithms (EAs) with the elite individual to participate in the evolution process. To illustrate the scalability of our proposal, we combine our proposal with the Genetic Algorithm (GA), Differential Evolution (DE), and CMA-ES. Experimental results in CEC2013 benchmark functions show our proposal has a broad prospect to estimate the elite individual and accelerate the convergence of optimization. 

\keywords{Evolutionary Algorithms (EAs), Proximity Optimal Principle (POP), Gaussian Process Regression (GPR), $\epsilon$-greedy acquisition function}
\end{abstract}
\section{Introduction} \label{sec:1}
Evolutionary Algorithms (EAs) generate the offspring by simulating natural behaviors, such as the selection, crossover, and mutation in Genetic Algorithm (GA)\cite{Sharma:14}, the movement of organisms in a bird flock or fish school in Particle Swarm Optimization (PSO)\cite{Ngatman:17}. Estimation of distribution algorithms (EDAs) generate the offspring by building the posterior probabilistic distribution model and sampling depending on certain strategies\cite{Zhang:05}. Both of them have achieved great success on complex optimization problems, such as combinatorial problems\cite{Gardner:13}, large-scale problems\cite{Siedlecki:89}, constraint problems\cite{Shen:17}, and multi-objective problems\cite{Hisao:96}.

However, conventional EAs are weak in learning the information of the fitness landscape and transforming the information into knowledge\cite{Zhou:15}, which means solutions generated by crossover and mutation operators may be close to the parents but far away from other promising solutions\cite{Zhang:05}. For EDAs, simple models are easy to implement but insufficient for complicated problems\cite{Pelikan:00}. A large population is required to build a good multivariate model, and the learning process (including network structure learning and parameter learning) can be very time-consuming\cite{Dong:13,Zhang:04}. Thus, an efficient EA should take advantage of both global distribution information and location information to produce promising solutions.

In addition, many studies\cite{Abdolmaleki:17,Fides:22} reveal the importance of finding trusting regions or elites in the fitness landscape, which can be fully applied to guide the direction of evolution well. Murata et al\cite{Noboru:15}. first proposed that a mathematical method could be used to calculate the global optimum using the information of two subsequent generations. Yu et al\cite{Yu:19}. proposed that in the differential evolution algorithm, the differential vector from the parent individual to its offspring (or the vector from an individual with poor fitness to an individual with higher fitness) is defined as the moving vector, and the convergence point is estimated according to the moving vector. A famous hypothesis, Proximity Optimal Principle (POP)\cite{Sadiq:13} states that well-performed solutions have a similar structure, and there is a high probability of better solutions existing around the elite individual. 

In this paper, we propose a novel method to estimate the elite individual in evolution inspired by Bayesian Optimization Algorithm\cite{Frazier:18}. In each generation of optimization, we build a posterior probabilistic model with Gaussian Process Regression (GPR) and estimate the elite individual by maximizing the $\epsilon$-greedy acquisition function\cite{De:19}. To illustrate the scalability of our proposal, we apply our proposal combined with Differential Evolution (DE)\cite{Chiang:13}, Genetic Algorithm (GA)\cite{Whitley:94}, and CMA-ES\cite{Auger:12}. Numerical experiments show that the participation of the elite individual can accelerate the convergence of optimization.

The remainder of this paper is organized as follows, Section \ref{sec:2} covers related works, including Bayesian Optimization Algorithm, acquisition functions, and the difference between Evolutionary Algorithms and Estimation of Distribution Algorithms. Section \ref{sec:3} introduces our proposal, hybridEAs. Section \ref{sec:4} shows the experiment results and analysis. Finally, Section \ref{sec:5} concludes this paper. 

\section{Related works} \label{sec:2}
\subsection{Bayesian Optimization Algorithm (BOA)} \label{sec:2.1}
BOA is a typical method of surrogate-assisted optimization. In practice, it has proved to be a very effective approach for single-objective expensive optimization problems with limited fitness evaluation times. Without loss of generality, the optimization problem can be formulated as 
\begin{equation}
	\label{eq:1}
	\begin{aligned}
		\max\limits_{x \in {\rm X}} f(x)
	\end{aligned}
\end{equation}
where ${\rm X} \subset \mathbb{R}^d$ is the feasible space and $f : \mathbb{R}^d \to \mathbb{R}$. Algorithm \ref{alg:1} outlines the procedure of standard BOA. 
\begin{algorithm}
	\label{alg:1}
	\caption{Bayesian Optimization Algorithm}
	\DontPrintSemicolon
	\SetAlgoLined
	\KwIn {${\rm Number \ of \ initial \ samples}:N;{\rm Maximum \ iteration}:M;{\rm Acquisition \ function}:a$}
	\KwOut {${\rm Best \ solution}: S$}
	\SetKwFunction{FBOA}{\textbf{BOA}}
	\SetKwProg{Fn}{Function}{:}{}
	\Fn{\FBOA{$N, M$}}{
		$X \gets \textbf{Sampling}(N)$ \;
		\For{$i=0 \ to \ N$}{
		    $F_i \gets f(X_i)$ \;
		}
		$S \gets \textbf{bestSample}(X, F)$ \;
		\For{$t=0 \ to \ M$}{
		    $m \gets \textbf{TrainGP}(X, F)$  \ \# Train a Gaussian Process model\; 
		    $x' \gets \textbf{argmax}_{x \in X} a(X, F, m)$ \ \# Maximize the acquisition function to determine the next sample point \;
		    $F' \gets f(x')$ \;
		    $S \gets \textbf{max}(F', S)$ \;
		    $X \gets X \cup x'$ \;
		    $F \gets F \cup F'$ \;
		}
		$\textbf{return} \ S$ 
	}
\end{algorithm}
BOA determines the coordinates of the next sample $x'$ by maximizing the acquisition function $a$. Up to now, many studies about the design of acquisition functions have been published, and various acquisition functions have different performances in balancing exploration and exploitation. 

\subsection{Acquisition functions} \label{sec:2.2}
The balance of exploration and exploitation is the most concerning issue for researchers in global optimization. As we described in Section \ref{sec:2.1}, different acquisition functions apply different strategies to balance exploration and exploitation. In this section, we will introduce 3 popular acquisition functions. Notice that all explanation is for the maximization problem.

\subsubsection{Probability of Improvement (PI)}
PI is one of the earliest proposed acquisition functions\cite{Kushner:64}, the basic idea is to maximize the probability of improvement between samples in the posterior probabilistic model and current best samples. Assuming the best-observed objective value $\tilde{f}$, PI applies the following strategy: 
\begin{equation}
	\label{eq:2}
	\begin{aligned}
		a_{PI}(x)=p(f>\tilde{f}|x, F)=\Phi(\frac{\mu (x)-\tilde{f}-\xi}{\sigma (x)})
	\end{aligned}
\end{equation}
$\mu(x)$ and $\sigma(x)$ are expectation and variance from the posterior probabilistic model. $\xi \ge 0$ is a hyperparameter. $\Phi(\cdot)$ is the cumulative distribution function of standard normal distribution. PI focuses on exploitation and easily guides the optimization trapped into the local optimum\cite{Donald:01}. Although $\xi$ allows PI certain randomness and the ability to get rid of local optimum, how determining a suitable $\xi$ is not easy work because PI is sensitive to $\xi$.

\subsubsection{Expected Improvement (EI)}
EI is derived from a thought experiment and becomes one of the most popular acquisition functions\cite{Jonas:75}. For the current best solution $\tilde{x}$ and its objective value $\tilde{f}$, if the objective value of next sample $\hat{x}$ is $\hat{f}$, then the improvement is 
\begin{equation}
	\label{eq:3}
	\begin{aligned}
		I(x, \tilde{f}, \hat{f}) = \max(\tilde{f} - \hat{f}, 0)
	\end{aligned}
\end{equation}
Then the expected improvement at $x$ may be expressed as\cite{Donald:98}
\begin{equation}
	\label{eq:4}
	\begin{aligned}
		a_{EI}(x)=\mathbb{E}(I(x, \tilde{f}))=\int_{-\infty}^{+\infty}I(x, \tilde{f}, \hat{f})p(\hat{f} |x, F)d\hat{f} \\
		=\sigma(x)(s\Phi(s)+\phi(s))
	\end{aligned}
\end{equation}
where $s=(\mu(x)-\tilde{f})/\sigma(x)$ is the predicted improvement at $x$ normalized by the uncertainty, and $\phi(\cdot)$ is the standard Gaussian probability density. Thus, EI balances the exploitation of solutions that are very likely to be a little better than $\tilde{f}$ with the exploration of others that may, with lower probability, turn out to be much better\cite{De:19}. Paper\cite{Bull:11} shows that under certain conditions, BOA with EI can achieve global optimum. 

\subsubsection{Upper Confidence Bound (UCB)}
As an optimization policy, UCB overestimates the mean with added uncertainty. Eq (\ref{eq:5}) shows the strategy of UCB.
\begin{equation}
	\label{eq:5}
	\begin{aligned}
		a_{UCB}(x)=\mu(x) + \sqrt{\beta_i}\sigma(x)
	\end{aligned}
\end{equation}
where $\beta_i \ge 0$ is the weight, which adaptively adjusts the importance of uncertainty by the number of evaluation times $i$. The addition of a multiple of the uncertainty means that the criterion prefers locations where the mean is large (exploitation) or the mean combined with the uncertainty is sufficiently large to warrant exploration\cite{De:19}.

\subsection{Evolutionary Algorithms vs Estimation of Distribution Algorithms} \label{sec:2.3}
\begin{figure}[htb]
	\centering
	\includegraphics[width=12cm]{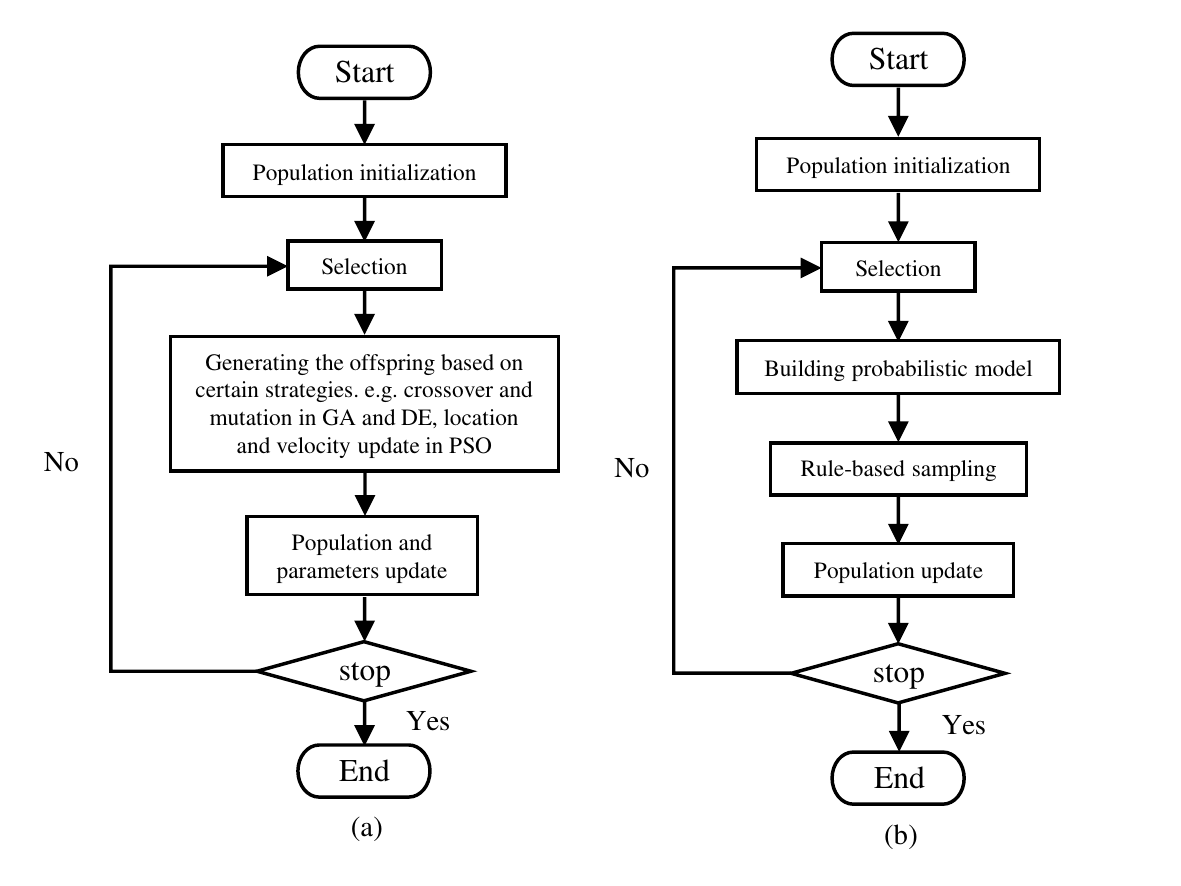}
	\caption{(a).The flowchart of EAs. (b).The flowchart of EDAs.}
	\label{fig:1}
\end{figure}

As we mentioned in Section \ref{sec:1}, EAs and EDAs generate offspring with different strategies. Fig. \ref{fig:1} demonstrates the main steps in EAs and EDAs. 

EAs simulate natural behaviors at the microscopic level. e.g. The crossover and mutation of the Genome in GA and DE. The velocity and position update in each dimension of PSO, etc. EAs focus on the contribution of each dimension to the objective value. On the other hand, EDAs lead the direction of optimization through probabilistic models, which means EDAs focus on the trust regions in the fitness landscape, and this macro-evolution is driven by objective values.

\section{hybridEAs} \label{sec:3}
\begin{figure}[htb]
	\centering
	\includegraphics[width=8cm]{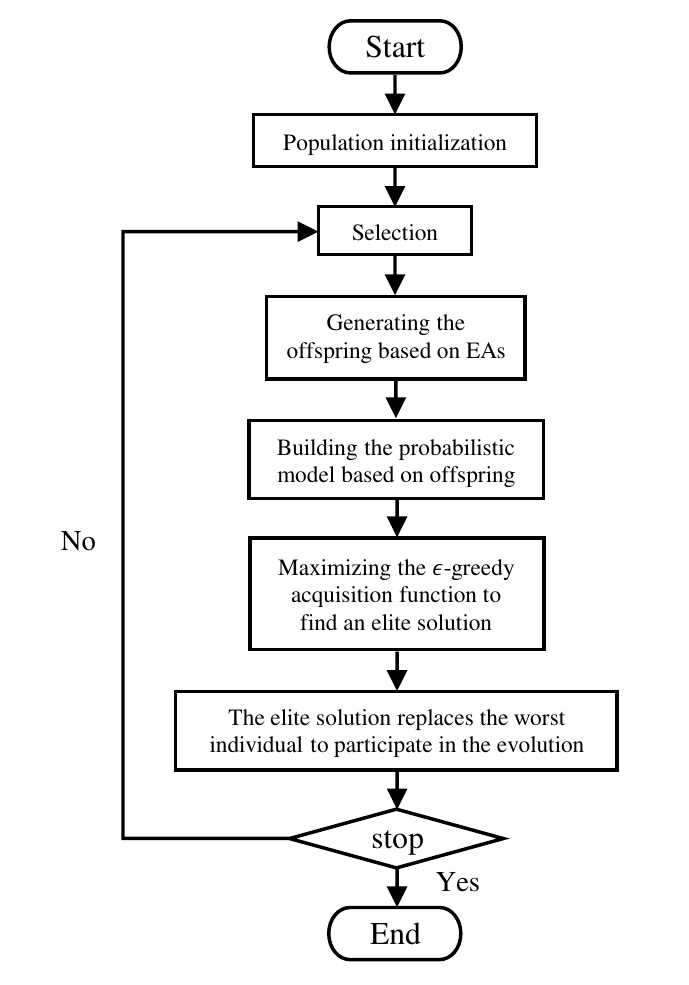}
	\caption{The flowchart of hybridEAs}
	\label{fig:2}
\end{figure}

In this section, we will introduce our proposal hybridEAs in detail. Here in Fig \ref{fig:2}, we demonstrate the flowchart of the main steps. 

After the initialization of the population, the algorithm selects the individuals to generate the offspring based on the strategies of EAs. Then, a probabilistic model is built based on offspring, $\epsilon$-greedy acquisition function guides the direction of optimization for searching an elite in the model. If the fitness of the elite is better than the worst individual in offspring, then the elite replaces this individual to participate in the optimization. 

\subsection{Offspring generated by EAs}
First, we follow the schemes of EAs to generate offspring. We apply DE, GA, and CMA-ES to verify the scalability of our proposal. Thus, after the crossover and mutation of GA, the mutation and crossover of DE, and random sampling in limited space of CMA-ES, the offspring is generated to construct the probabilistic model.

\subsection{Elite found by probabilistic model}
We apply GPR to construct a probabilistic model, a comprehensive introduction can be found in \cite{Rasmussen:03}. In essence, a GP is a collection of random variables, and any finite number of these have a joint Gaussian distribution. 

As an acquisition function, the $\epsilon$-greedy scheme is proposed in \cite{De:19} motivated by the successful implementation in reinforcement learning\cite{Mnih:15}. Considering the balance of exploration and exploitation, the greedy strategy mostly selects the exploitative solution, and $\epsilon$ allows the exploration behaviors to be adopted in the optimization. This pseudocode of $\epsilon$-greedy scheme is shown in Algorithm \ref{alg:2}.
\begin{algorithm}
	\label{alg:2}
	\caption{$\epsilon$-greedy scheme}
	\DontPrintSemicolon
	\SetAlgoLined
	\KwOut {${\rm next \ sample \ position}: x'$}
	\SetKwFunction{FG}{\textbf{$\epsilon$-greedy}}
	\SetKwProg{Fn}{Function}{:}{}
	\Fn{\FG{}}{
		\If{$\textbf{rand}()<\epsilon$} {
		    $x' \gets \textbf{randChoice}(X)$ \;
		}
		\Else {
		    $x' \gets \textbf{argmax}_{x \in X}(\mu(x))$ \;
		}
		$\textbf{return} \ x'$ 
	}
\end{algorithm}
$\epsilon$ is a hyperparameter to control the proportion of exploration and exploitation. Sensitivity test in \cite{De:19} prove the performance of optimization is not sensitive to the precise value of $\epsilon$. As the dimension of the problem increases, the need for exploration via deliberate inclusion of exploratory moves turns out to be less important, one of the reasons is due to the low fidelity of the surrogate model.

\section{Numerical Experiment and Analysis} \label{sec:4}
In this Section, experiments are implemented to evaluate our proposal. Section \ref{sec:4.1} introduces the experiment settings. Section \ref{sec:4.2} shows the experimental results. Finally, the analysis is provided in Section \ref{sec:4.3}.

\subsection{Experiment settings} \label{sec:4.1}
Here, we define shortened names of our proposal:

(a) hGA: our proposal combined with GA, the shortened name of hybrid GA. 

(b) hGA: our proposal combined with DE, the shortened name of hybrid DE. 

(b) hCMA-ES: our proposal combined with CMA-ES, the shortened name of hybrid CMA-ES. 

We apply 28 benchmark functions from the CEC2013 Suite\cite{Liang:13} with independent 30 trial runs in this evaluation experiment. The dimensions of functions are $[2, 5, 10]$-D. Population size and maximum evaluation times are set to $50 \times $D and $1000 \times $D respectively. $\epsilon$ in $\epsilon$-greedy acquisition function is set to 0.1. At the end of optimization, we apply the Mann-Whitney U test to the fitness values. If hybridEAs are significantly better than the original EAs, we apply $\gg$ (p$<$0.01) and $>$ (p$<$0.05) to denote the significance, $\approx$ represents there is no significance between two comparing methods.

\subsection{Experimental results} \label{sec:4.2}
Table \ref{tbl:1} shows the statistical results between hybrid EAs and EAs.
\begin{sidewaystable}[tbh]
	\scriptsize
	\centering
	\caption{Experimental results on CEC2013 Suite}
	\label{tbl:1}
	\begin{tabular}{cccccccccc}
		\toprule
		\multirow{2}{*}{Func.} & \multicolumn{3}{c}{GA ($Cr=0.5, M=0.1$)} &  \multicolumn{3}{c}{DE ($F=0.7, Cr=0.9$)} & \multicolumn{3}{c}{CMA-ES ($\sigma=1.3$)}  \\
		\cmidrule(r){2-4} \cmidrule(r){5-7} \cmidrule(r){8-10}
		~ & 2-D & 5-D & 10-D & 2-D & 5-D & 10-D & 2-D & 5-D & 10-D\\
		\midrule 
		$f_1$ & hGA $\gg$ GA& hGA $\gg$ GA & hGA $>$ GA & hDE $\gg$ DE & hDE $\gg$ DE & hDE $\gg$ DE & hCMA-ES $\gg$ CMA-ES & hCMA-ES $\gg$ CMA-ES & hCMA-ES $\gg$ CMA-ES \\
		$f_2$ & hGA $\gg$ GA& hGA $\gg$ GA & hGA $\gg$ GA & hDE $\gg$ DE & hDE $\gg$ DE & hDE $\gg$ DE & hCMA-ES $\gg$ CMA-ES & hCMA-ES $\gg$ CMA-ES & hCMA-ES $\gg$ CMA-ES \\
		$f_3$ & hGA $\gg$ GA& hGA $\gg$ GA & hGA $\gg$ GA & hDE $\gg$ DE & hDE $\gg$ DE & hDE $\gg$ DE & hCMA-ES $\gg$ CMA-ES & hCMA-ES $\gg$ CMA-ES & hCMA-ES $\gg$ CMA-ES \\
		$f_4$ & hGA $\approx$ GA& hGA $\approx$ GA & hGA $\approx$ GA & hDE $\gg$ DE & hDE $\gg$ DE & hDE $\gg$ DE & hCMA-ES $\gg$ CMA-ES & hCMA-ES $\approx$ CMA-ES & hCMA-ES $\approx$ CMA-ES \\
		$f_5$ & hGA $\gg$ GA& hGA $\gg$ GA & hGA $\gg$ GA & hDE $\gg$ DE & hDE $>$ DE & hDE $\gg$ DE & hCMA-ES $\gg$ CMA-ES & hCMA-ES $\gg$ CMA-ES & hCMA-ES $\gg$ CMA-ES \\
		$f_6$ & hGA $\gg$ GA& hGA $\gg$ GA & hGA $\gg$ GA & hDE $\gg$ DE & hDE $\gg$ DE & hDE $\approx$ DE & hCMA-ES $\gg$ CMA-ES & hCMA-ES $\gg$ CMA-ES & hCMA-ES $\gg$ CMA-ES \\
		$f_7$ & hGA $>$ GA& hGA $\approx$ GA & hGA $\approx$ GA & hDE $>$ DE & hDE $\approx$ DE & hDE $\approx$ DE & hCMA-ES $\gg$ CMA-ES & hCMA-ES $\approx$ CMA-ES & hCMA-ES $\approx$ CMA-ES \\
		$f_8$ & hGA $\approx$ GA& hGA $\approx$ GA & hGA $\approx$ GA & hDE $\gg$ DE & hDE $\approx$ DE & hDE $\gg$ DE & hCMA-ES $\approx$ CMA-ES & hCMA-ES $\approx$ CMA-ES & hCMA-ES $\approx$ CMA-ES \\
		$f_9$ & hGA $\approx$ GA& hGA $\approx$ GA & hGA $\approx$ GA & hDE $\gg$ DE & hDE $\approx$ DE & hDE $\approx$ DE & hCMA-ES $\gg$ CMA-ES & hCMA-ES $\approx$ CMA-ES & hCMA-ES $\approx$ CMA-ES \\
		$f_{10}$ & hGA $\gg$ GA& hGA $\approx$ GA & hGA $\approx$ GA & hDE $\gg$ DE & hDE $\gg$ DE & hDE $\gg$ DE & hCMA-ES $\approx$ CMA-ES & hCMA-ES $\approx$ CMA-ES & hCMA-ES $\approx$ CMA-ES \\
		$f_{11}$ & hGA $\gg$ GA& hGA $\gg$ GA & hGA $\approx$ GA & hDE $\gg$ DE & hDE $\gg$ DE & hDE $\approx$ DE & hCMA-ES $\approx$ CMA-ES & hCMA-ES $\approx$ CMA-ES & hCMA-ES $\approx$ CMA-ES \\
		$f_{12}$ & hGA $\gg$ GA& hGA $\approx$ GA & hGA $\approx$ GA & hDE $\gg$ DE & hDE $\approx$ DE & hDE $\approx$ DE & hCMA-ES $\gg$ CMA-ES & hCMA-ES $>$ CMA-ES & hCMA-ES $\approx$ CMA-ES \\
		$f_{13}$ & hGA $>$ GA& hGA $\approx$ GA & hGA $\approx$ GA & hDE $\gg$ DE & hDE $\gg$ DE & hDE $\approx$ DE & hCMA-ES $\approx$ CMA-ES & hCMA-ES $\approx$ CMA-ES & hCMA-ES $\approx$ CMA-ES \\
		$f_{14}$ & hGA $\gg$ GA& hGA $\approx$ GA & hGA $\approx$ GA & hDE $\gg$ DE & hDE $\approx$ DE & hDE $\approx$ DE & hCMA-ES $\gg$ CMA-ES & hCMA-ES $\approx$ CMA-ES & hCMA-ES $\approx$ CMA-ES \\
		$f_{15}$ & hGA $\gg$ GA& hGA $\approx$ GA & hGA $\approx$ GA & hDE $\gg$ DE & hDE $\approx$ DE & hDE $\approx$ DE & hCMA-ES $\approx$ CMA-ES & hCMA-ES $\approx$ CMA-ES & hCMA-ES $\approx$ CMA-ES \\
		$f_{16}$ & hGA $\approx$ GA& hGA $\approx$ GA & hGA $\approx$ GA & hDE $\approx$ DE & hDE $\approx$ DE & hDE $\approx$ DE & hCMA-ES $\gg$ CMA-ES & hCMA-ES $\gg$ CMA-ES & hCMA-ES $\approx$ CMA-ES \\
		$f_{17}$ & hGA $\gg$ GA& hGA $\approx$ GA & hGA $\approx$ GA & hDE $\approx$ DE & hDE $\approx$ DE & hDE $\approx$ DE & hCMA-ES $\approx$ CMA-ES & hCMA-ES $\approx$ CMA-ES & hCMA-ES $\approx$ CMA-ES \\
		$f_{18}$ & hGA $\approx$ GA& hGA $\approx$ GA & hGA $\gg$ GA & hDE $\approx$ DE & hDE $\approx$ DE & hDE $\approx$ DE & hCMA-ES $>$ CMA-ES & hCMA-ES $\approx$ CMA-ES & hCMA-ES $\approx$ CMA-ES \\
		$f_{19}$ & hGA $\gg$ GA& hGA $\approx$ GA & hGA $\approx$ GA & hDE $\gg$ DE & hDE $\approx$ DE & hDE $\approx$ DE & hCMA-ES $\gg$ CMA-ES & hCMA-ES $\approx$ CMA-ES & hCMA-ES $\approx$ CMA-ES \\
		$f_{20}$ & hGA $\gg$ GA& hGA $\approx$ GA & hGA $\approx$ GA & hDE $>$ DE & hDE $\approx$ DE & hDE $\approx$ DE & hCMA-ES $\approx$ CMA-ES & hCMA-ES $\approx$ CMA-ES & hCMA-ES $\approx$ CMA-ES \\
		$f_{21}$ & hGA $\gg$ GA& hGA $>$ GA & hGA $\approx$ GA & hDE $\gg$ DE & hDE $\gg$ DE & hDE $\gg$ DE & hCMA-ES $\gg$ CMA-ES & hCMA-ES $\gg$ CMA-ES & hCMA-ES $\gg$ CMA-ES \\
		$f_{22}$ & hGA $\approx$ GA& hGA $\approx$ GA & hGA $\gg$ GA & hDE $\approx$ DE & hDE $\gg$ DE & hDE $>$ DE & hCMA-ES $\gg$ CMA-ES & hCMA-ES $\approx$ CMA-ES & hCMA-ES $\approx$ CMA-ES \\
		$f_{23}$ & hGA $\approx$ GA& hGA $\approx$ GA & hGA $\approx$ GA & hDE $\gg$ DE & hDE $\approx$ DE & hDE $\approx$ DE & hCMA-ES $\approx$ CMA-ES & hCMA-ES $\approx$ CMA-ES & hCMA-ES $\approx$ CMA-ES \\
		$f_{24}$ & hGA $\gg$ GA& hGA $\approx$ GA & hGA $\approx$ GA & hDE $\gg$ DE & hDE $\approx$ DE & hDE $\approx$ DE & hCMA-ES $>$ CMA-ES & hCMA-ES $\approx$ CMA-ES & hCMA-ES $\approx$ CMA-ES \\
		$f_{25}$ & hGA $\gg$ GA& hGA $\approx$ GA & hGA $\approx$ GA & hDE $\approx$ DE & hDE $\approx$ DE & hDE $\approx$ DE & hCMA-ES $>$ CMA-ES & hCMA-ES $\approx$ CMA-ES & hCMA-ES $\approx$ CMA-ES \\
		$f_{26}$ & hGA $\gg$ GA& hGA $\approx$ GA & hGA $\approx$ GA & hDE $\gg$ DE & hDE $\approx$ DE & hDE $\approx$ DE & hCMA-ES $\approx$ CMA-ES & hCMA-ES $\approx$ CMA-ES & hCMA-ES $\approx$ CMA-ES \\
		$f_{27}$ & hGA $\approx$ GA& hGA $\approx$ GA & hGA $\approx$ GA & hDE $\approx$ DE & hDE $\approx$ DE & hDE $\approx$ DE & hCMA-ES $>$ CMA-ES & hCMA-ES $\approx$ CMA-ES & hCMA-ES $\gg$ CMA-ES \\
		$f_{28}$ & hGA $\approx$ GA& hGA $\approx$ GA & hGA $\approx$ GA & hDE $\gg$ DE & hDE $\approx$ DE & hDE $\approx$ DE & hCMA-ES $\gg$ CMA-ES & hCMA-ES $\gg$ CMA-ES & hCMA-ES $\gg$ CMA-ES \\
		\bottomrule
	\end{tabular}
\end{sidewaystable}
\subsection{Analysis} \label{sec:4.3}
From the experiment results, the participation of elite individuals can accelerate the convergence of optimization in unimodal functions ($f_1-f_5$) significantly, which means the probabilistic model is efficient and accurate, and the elite individuals guide the optimization to the correct direction. However, as the complexity of the fitness landscape and the dimension of decision variables increase, the computational cost increases, and the accuracy of the surrogate model also degrades rapidly. Although the $\epsilon$-greedy acquisition function can avoid the misleading optimization of wrong elite individuals to a certain extent, the decrease in the accuracy of the surrogate model is the main reason that our proposal did not well-performed as expected on the multimodal ($f_6-f_{20}$) and composition ($f_{21}-f_{28}$) problem. 

\section{Conclusion} \label{sec:5}
This paper proposes a method to estimate the elite individual by GPR and $\epsilon$-greedy acquisition function. Numerical experiments show our proposal can improve the performance of conventional EAs and has strong scalability to combine with various EAs. Finally, we list some open topics which may improve the performance of our proposal:

\subsection{The employment of dimension reduction techniques}
Although the increase in computational cost and degeneration of model accuracy make our proposal limited in high-dimensional problems, the dimension reduction techniques may alleviate this issue. However, the information loss in the procedure of dimension reduction and restoration will also affect the accuracy of estimation. How to apply dimension reduction techniques under suitable conditions is a promising topic.

\subsection{Sampling around promising individuals}
As the statement of POP, well-performed individuals have a similar structure. Thus, the implementation of sampling techniques or local search operators around promising individuals may find superior individuals, and the participation of these superior individuals will accelerate the convergence of optimization. 

 \section{Acknowledgement}
This work was supported by JSPS KAKENHI Grant Number JP20K11967.
%
%
\bibliographystyle{splncs04}
\bibliography{paper}
\end{document}